\documentclass[letterpaper, 10 pt, conference]{ieeeconf}  

\IEEEoverridecommandlockouts                              

\usepackage[dvipdfmx]{graphicx} 
\usepackage{multirow}
\usepackage{arydshln}
\usepackage{url}
\usepackage{spverbatim}
\usepackage{ascmac}
\usepackage{algorithm}
\usepackage{overpic}
\usepackage[noend]{algorithmic}
\usepackage{amsmath}

\usepackage{tcolorbox}
\tcbuselibrary{breakable} 

\title{\LARGE \bf
Proprioception Enhances Vision Language Model \\in Generating Captions and Subtask Segmentations for Robot Task
}

\author{
Kanata Suzuki$^{1,2}$, 
Shota Shimizu$^{1}$, 
and Tetsuya Ogata$^{1,3}$
\thanks{
All authors are affiliated with Faculty of Science and Engineering, Waseda University, Tokyo 169-8050, Japan. 
$^{1,2}$Kanata Suzuki is also at Artificial Intelligence Laboratory, Fujitsu Limited, Kanagawa 211-8588, Japan. 
$^{1,3}$Tetsuya Ogata is also at the National Institute of Advanced Industrial Science and Technology, Tokyo 100-8921, Japan. 
E-mail:{\tt\small suzuki.kanata@fujitsu.com}
}}

\begin{document}

\maketitle
\thispagestyle{empty}
\pagestyle{empty}


\begin{abstract}

From the perspective of future developments in robotics, it is crucial to verify whether foundation models trained exclusively on offline data, such as images and language, can understand the robot motion.
In particular, since Vision Language Models (VLMs) do not include low-level motion information from robots in their training datasets, video understanding including trajectory information remains a significant challenge.
In this study, we assess two capabilities of VLMs through a video captioning task with low-level robot motion information: (1) automatic captioning of robot tasks and (2) segmentation of a series of tasks. Both capabilities are expected to enhance the efficiency of robot imitation learning by linking language and motion and serve as a measure of the foundation model's performance.
The proposed method generates multiple ``scene" captions using image captions and trajectory data from robot tasks.
The full task caption is then generated by summarizing these individual captions. 
Additionally, the method performs subtask segmentation by comparing the similarity between text embeddings of image captions.
In both captioning tasks, the proposed method aims to improve performance by providing the robot's motion data — joint and end-effector states — as input to the VLM.
Simulator experiments were conducted to validate the effectiveness of the proposed method.

\end{abstract}

\section{INTRODUCTION}
\label{sec1}

In recent years, Large Language Models (LLMs), Vision Language Models (VLMs), and other foundation models have been applied to planning and recognition tasks in robotics, achieving notable results~\cite{CLIPort}\cite{Ahn2022a}\cite{kirillov2023segany}.
However, it remains unclear whether these models truly understand the dynamics of robot motion.
Most foundation models do not include the robot's proprioception, such as sensorimotor trajectories, in their training datasets.
Robot motion information associated with the body is extremely important for building a world model that includes physical dynamics.
Due to this lack of that input, previous studies on VLMs and LLMs handle motion trajectory outputs in predefined API formats~\cite{Liang2022a}, which are often described as superficial~\cite{alice2024}\cite{mollo2023}\cite{harnad2025}.
Therefore, investigating the robot motion understanding capabilities of VLMs is a crucial research topic for advancing robot foundation models~\cite{vla}\cite{rtx}\cite{gr00t}\cite{go1}\cite{pi05}.

One way to measure a VLM's understanding of the robot's motion is through video captioning~\cite{yoshino}.
The skills necessary for this task can be categorized into two types:
\begin{enumerate}
    \item Automatic captioning of the robot task
    \item Segmentation of a series of motions
\end{enumerate}
The first skill is the ability to describe a series of robot motions in natural language when provided with task-related images or videos.
The language-motion pair data generated through captioning is crucial for integrated learning that connects robot actions with linguistic instructions~\cite{jang2021bc}\cite{Toyoda2022}\cite{Shi2024}\cite{Suzuki2024}.

Moreover, for a robot to perform tasks effectively in real environments, it requires specific, executable motion instructions rather than mere descriptions of its actions.
The second skill is the ability to divide a series of robot tasks into multiple subtasks. Previous studies have proposed methods in which subtask divisions are either predefined by humans~\cite{Fujii2022sii} or learned through training~\cite{beal}\cite{nakajo}.
If VLMs can extract subtasks from complex and long-horizon tasks in a zero-shot manner, more efficient robot learning is likely to be achieved.

\begin{figure}[tb]
    \centering
    \includegraphics[width=\columnwidth]{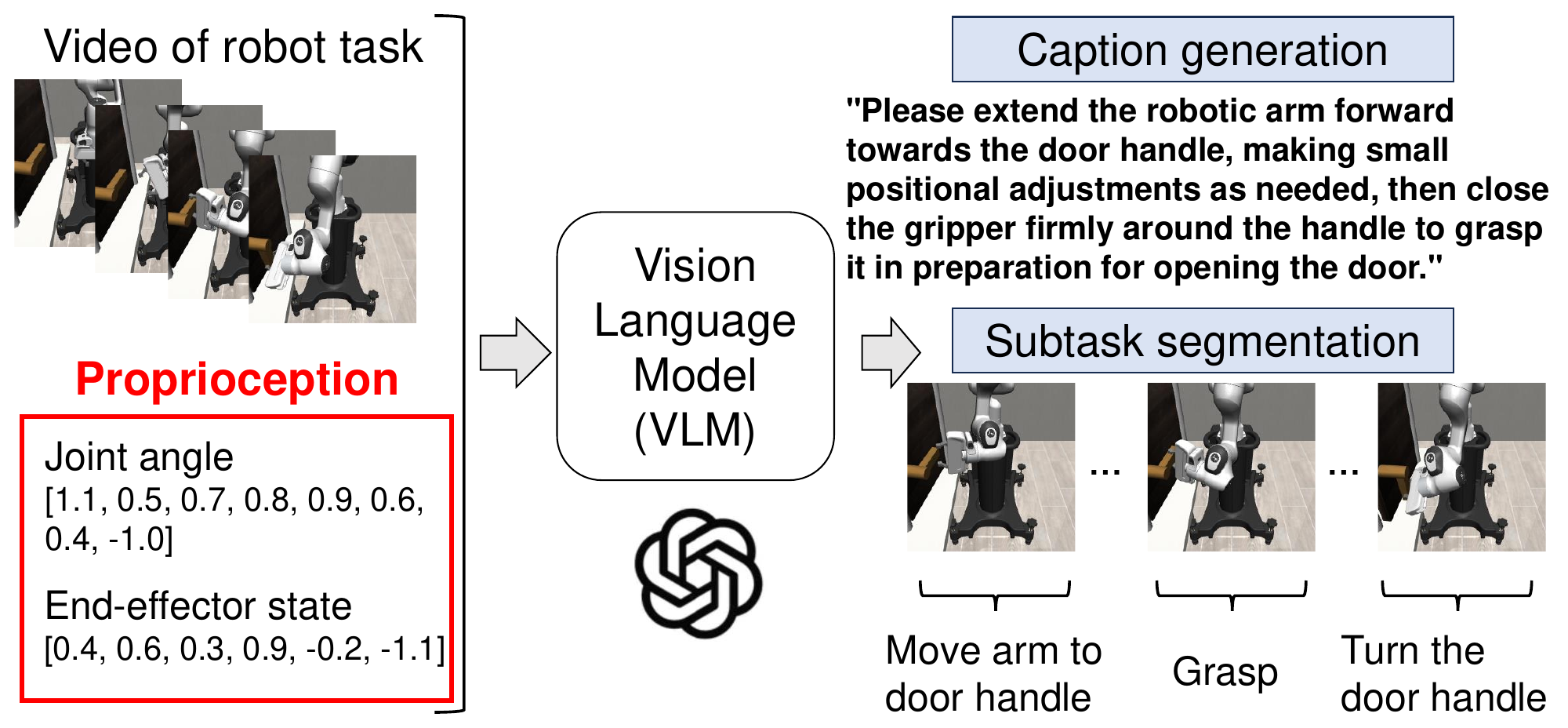}
    \caption{Overview of this study. We verified the performance of VLM by generating captions and subtask segmentations of robot motions.}
    \label{fig1}
\end{figure}

In this study, we propose a method for robot motion captioning and subtask segmentation using a VLM.
We believe that proprioception is important for understanding robot actions, and we take the approach of incorporating it into the VLM prompts.
Through the two captioning tasks, we aim to verify whether the VLM can comprehend robot motion and generate outputs that account for the robot's trajectory (Fig.~\ref{fig1}).
The proposed method builds on a video caption generation approach~\cite{amirian}, generating ``scene" captions from evenly spaced image captions and summarizing them into a complete task description.
We also segment a series of motions into subtasks by comparing the text similarity between image captions.
Additionally, the method enhances caption accuracy by incorporating low-level robot trajectory data (joint and end-effector states paired with images) into the prompt that is input to the VLM.
In the experiment, we validate the effectiveness of the proposed method by applying it to several robot tasks recorded through human operation in a simulator.

\section{RELATED WORK}
\label{sec2}

\subsection{Video Captioning}
In image processing, video captioning is widely studied as a means of linking time-series data with natural language~\cite{amirian}\cite{Vid2Seq}.
The primary goal in this field is to generate titles and summaries for movies and videos.
Amirian et al.~\cite{amirian} generated sub-captions for representative image frames and then summarized them to produce a title and abstract for the video.
Yang et al.~\cite{Vid2Seq} enhanced a language model with time tokens to enable seamless caption prediction.
They defined the end of a sentence in transcribed audio as a pseudo-event boundary to align captions with corresponding video segments.
Although these studies achieve high-accuracy caption generation, most previous work did not focus on robot motions.
Additionally, the input data typically lacked key information, such as the motion trajectory.

\subsection{Segmentation of Robot Motion}
In contrast, many studies have focused on segmenting robot motions using machine learning~\cite{yoshino}\cite{beal}\cite{nakajo}\cite{nagano}.
Beal et al.~\cite{beal} proposed a motion segmentation method based on a Hidden Markov Model (HMM) that accounts for infinite time states.
Nagano et al.~\cite{nagano} extended the HMM by integrating a variational autoencoder, enabling motion segmentation that considers the duration of each segment.
Yoshino et al.~\cite{yoshino} introduced a method that embeds actuator and camera transitions into a model through unsupervised learning, clustering motion classes using the $k$-nearest neighbor method.
Nakajo et al.~\cite{nakajo} incorporated parametric bias (PB) into an imitation learning framework, proposing an approximate subtask segmentation method by grouping PBs with similar values.
While these studies rely on the learning trajectory of the robot's task experience, they do not address the performance evaluation of trained VLMs, which is the focus of this study.
The robot tasks segmented in the studies mentioned above can be incorporated into an imitation learning framework as individual training data. It has also been suggested that subtasks could be embedded into a model in a manner that allows them to be switched and combined~\cite{Kase2018icra}. Thus, achieving automatic subtask segmentation has the potential to significantly enhance the performance of robot learning.

\subsection{VLA and World Model}
In recent years, there has been active progress in developing the Vision Language Action Model (VLA~\cite{vla}\cite{rtx}\cite{gr00t}\cite{go1}\cite{pi05}), 
The VLA aims to create a foundation model that incorporates physical dynamics by fine-tuning the VLM to predict robot motion information. 
Recent advances in VLA have shown that pre-training a VLM to predict latent actions learned from video data can improve performance~\cite{lapa}.
However, this does not measure the performance of the original VLM.

Additionally, the world foundation model leverages a VLM trained on over 20 million hours of video data, enabling video output that reflects physical dynamics~\cite{cosmos}. 
Some studies have incorporated world models into VLA to connect videos of human tasks with robot behavior~\cite{unipi}\cite{gr-1}.
As VLM's ability to understand robot behavior improves, it is expected that the prediction accuracy of these studies will also improve.

And, in \cite{alter3}, the input prompt for the LLM is modified to include joint angle information, allowing it to directly output the robot's trajectory. 
This suggests that foundation models may have the potential to predict robot dynamics through prompt engineering.
In this study, we use the zero-shot inference capabilities of VLM to perform captioning of robot motions.
Our contributions can be summarized as follows:
\begin{itemize}
    \item Proposal of a motion captioning and segmentation method using VLM.
    \item Verification and consideration of changes in VLM performance by including robot motion information.
\end{itemize}

\section{PROPOSED METHOD}
\label{sec3}

\begin{figure*}[tb]
    \centering
    \includegraphics[width=2.0\columnwidth]{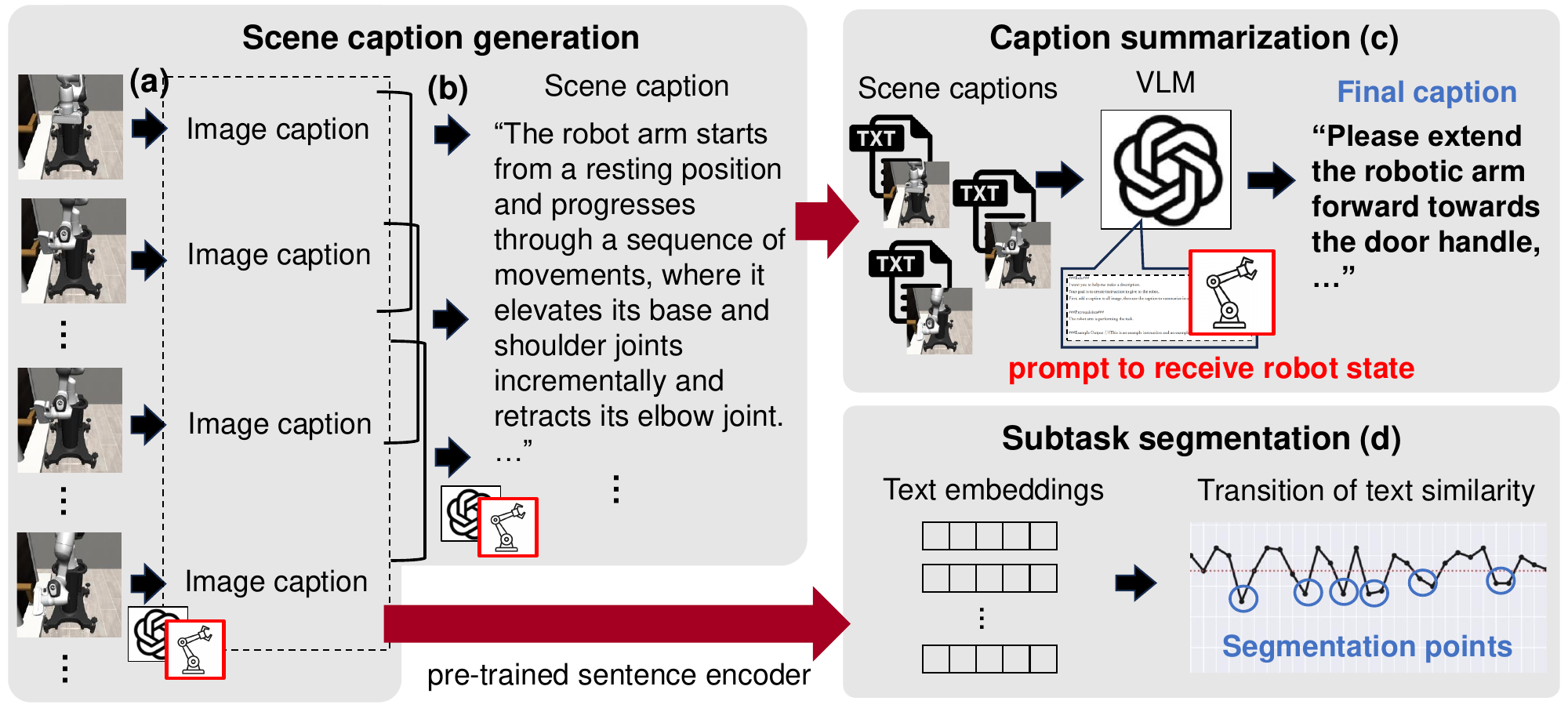}
    \caption{Overview of the proposed method.}
    \label{fig2}
\end{figure*}

An overview of the proposed method is shown in Fig.~\ref{fig2}.
Our proposed method is based on a previous study of video captioning~\cite{amirian}.
We generate a final caption of the entire robot task by performing step-by-step caption generation (Section~\ref{sec3_1}).
The proposed method enhances caption accuracy by incorporating low-level robot trajectory data into the prompt that is input to the VLM.
Then, we automatically divide the robot task by measuring the similarity of the image captions generated during the captioning process (Section~\ref{sec3_2}).

\subsection{Caption Generation by VLM with Robot State}
\label{sec3_1}
\subsubsection{Scene Caption Generation}
First, we describe the procedure for generating captions for robot motions. The robot motion sequence $D=((i_0,a_0), \cdots, (i_n,a_n)))$ used in this study consists of pairs of motion information and images extracted every 20 frames from the original motion sequence.
Here, $i$ represents the camera image and $a$ represents the robot trajectory data, which includes joint angles and the end-effector's posture and position.
For all images in $D$, image captions $(d_0, \cdots, d_n)$ are generated by the VLM (Fig.~\ref{fig2}a).
After that, the generated image captions are input to the VLM in groups of $k$ to obtain a set of scene captions $(d'_0, \cdots, d'_{n/k})$ (Fig.~\ref{fig2}b).
In this study, we set $k=5$.
The scene caption represents a short robot motion. By inputting multiple frames together, the VLM output remains stable despite minor variations in the image. Steps (a–b) are achieved by providing the VLM with the image and corresponding robot motion information, along with the prompt shown below.

\begin{tcolorbox}[breakable=true,boxrule=0mm, colback = yellow!25!white, sharp corners]
\#\#\# Task \#\#\#\\
Your task is to create instructions to give to the robot. \\
Input is a sequence of robot motion frame images, robot joint angles, and robot end-effector position information. \\
Output is a caption to each image based on what the robot arm is doing. \\
\\
\#\#\# Joint/Pose \#\#\#\\
The robot arm has seven degrees of freedom. \\
The joint angle information consists of eight key elements: seven joint angles and one representing the opening of the robot gripper. For example, in the sequence [1.1, 0.5, 0.7, 0.8, 0.9, 0.6, 0.4, -1.0], the first seven values (1.1 to 0.4) correspond to the joint angles, while the last value (-1.0) indicates the gripper's opening state. \\
There are six pieces of information for fingertip position and posture. \\
For example, in the case of [0.4, 0.6, 0.3, 0.9, 0.2, -1.1], 0.4, 0.6, 0.3 represent x, y, z coordinates, which are position, and 0.9, 0.2, -1.1 represent roll, pitch, and yaw, which are orientation. \\
\\
\#\#\# Guidelines \#\#\#\\
1. Describe the caption simply. \\
2. The robot arm is doing its task. \\
3. In the caption field, please provide "caption" only. \\
4. When creating captions, summaries, or generating instructions, be sure to refer to the joint angles and robot hand position information. \\
\\
\# Process 1\\
Describe the robot arm's motion at each step based on a series of images and the corresponding joint angle and hand position data. Each series consists of five images.
\end{tcolorbox}

\begin{tcolorbox}[breakable=true,boxrule=0mm, colback = yellow!25!white, sharp corners]
\# Process 2\\
Based on each of the described robotic arm actions, summarize the overall work situation. This summary should aim to provide an understanding of the purpose of the robotic arm and the sequence of tasks it is performing.
\end{tcolorbox}

In the captioning phase, the VLM is fed with image and text pairs. 
The text contains the robot motion information (joint angles and the end-effector's posture and position).
Due to space constraints, the concrete example of the cube-stacking task is omitted, but it is included in the prompt.
This set of scene captions serves as input for the VLM during the summarization process, which is explained next.

\subsubsection{Caption Summarization}
Next, all the scene captions $(d'_0, \cdots, d'_{n/k})$ generated in the previous step are summarized to create a caption for the entire robot motion (Fig.~\ref{fig2}c).
The images and motion data used to generate each scene caption are reintroduced as input to the VLM.
By summarizing the scene captions, it becomes possible to generate a comprehensive caption that reflects the overall flow, even for long robot tasks.
This process is guided by the prompt shown below.

\begin{tcolorbox}[breakable=true,boxrule=0mm, colback = yellow!25!white, sharp corners]
\# Process 3\\
Based on the summary of the task, generate a single, specific instruction for the robotic arm. This directive should clearly and precisely describe the action required to complete the task. Keep the instruction brief and focused, incorporating joint angle and hand position information.
\end{tcolorbox}

Here, a concrete example of the cube stacking task is also included. We initially experimented with a method that excluded image and motion information during the summarization phase. However, since the output lacked stability, we adopted a prompt that incorporates the robot's state. In this study, the prompt is designed to generate final captions as annotations for robot movements, but it can be adapted flexibly based on the intended application.

\subsection{Subtask Segmentation}
\label{sec3_2}
Finally, the entire task is divided into subtasks based on the image captions obtained in the captioning process (Fig.~\ref{fig2}d).
We input all image captions into a pre-trained sentence encoder and obtain embedding vectors $(e_0, \cdots, e_n)$.
By calculating the cosine similarity between each embedding vector $e_j$ and the embedding vector $e_{j+1}$ one step later, we determine how much description difference exists between corresponding motion segments.
If the obtained cosine similarity values are below a threshold, we consider them to be different subtasks (segmentation points).
Experiments were conducted to investigate the behavior of VLM by evaluating the number of segmentation points by changing this threshold value.

\section{EXPERIMENTS}
\label{sec4}

\begin{figure}[tb]
    \centering
    \includegraphics[width=\columnwidth]{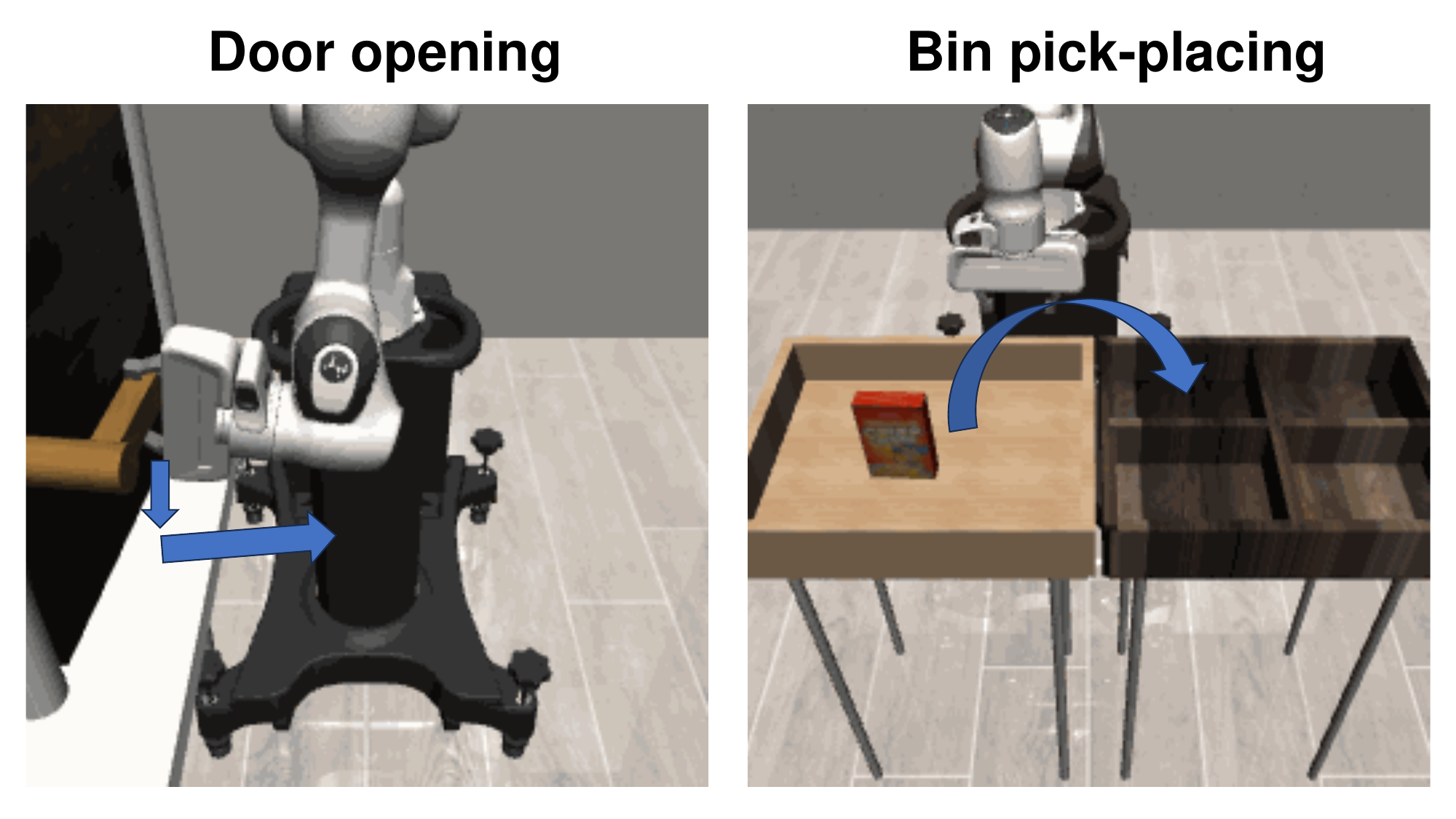}
    \caption{Experimental setup. We verified our method in the Door opening and Bin pick-placing task.}
    \label{fig3}
\end{figure}

\begin{figure*}[t]
    \centering
    \includegraphics[width=2.0\columnwidth]{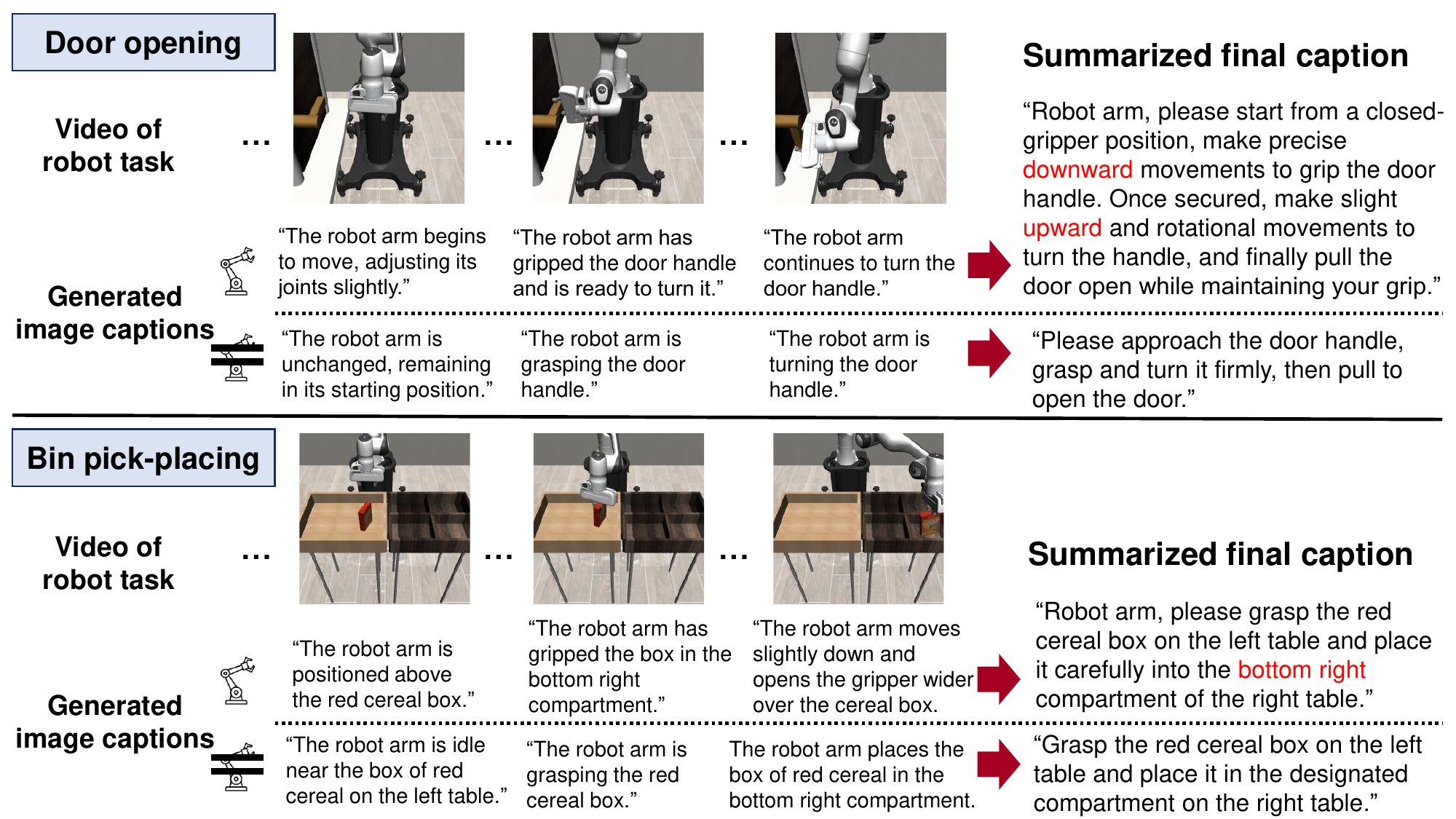}
    \caption{Examples of generated sub-captions and summarized caption.}
    \label{fig4}
\end{figure*}


We conducted experiments to generate captions for some robot tasks recorded on a simulator and verify the effectiveness of the proposed method.
We targeted two tasks implemented in the Robosuite simulator~\cite{robosuite2020}: Door opening and Bin pick-placing for cereal (Fig.~\ref{fig3}).
As robot motion data, we recorded 22 sequences operated by a human using a 3D mouse.
10 sequences are for Door opening and 12 sequences are for Bin pick-placing (3 sequences for each bin position).
The motion data included RGB images of 256$\times$256 pixels, 8 degree-of-freedom joint angles including gripper opening and closing, and the end-effector's posture.
The camera was positioned so that the images could capture the entire robot task.
The sequence length for each task was 580 and 539 steps on average for Door opening and Bin pick-placing, respectively.

In the caption generation task, we evaluated whether the final captions accurately described key elements of each task, including the object being manipulated (e.g., doorknob, cereal), the action (e.g., opening, pick-and-place), and trajectory information (e.g., motion direction, bin position). If all these elements were correctly captured, we determined that the VLM successfully generated captions containing detailed motion trajectory information. In the subtask segmentation task, we converted the image captions generated by the VLM into 768-dimensional embeddings using the stsb-xlm-r-multilingual model from the sentence-transformers library and calculated the cosine similarity between consecutive image captions. We examined how the subtask segmentation behavior of the VLM varied based on the similarity threshold between captions. For our experiments, we used ChatGPT-4V~\cite{ChatGPT} as the VLM, setting the temperature to zero for all tests. ChatGPT is one of the most widely used LLM/VLM models in academic research~\cite{Vemprala2023}.
 The version employed in our study was gpt-4v-preview.

\section{RESULTS AND DISCUSSION}
\label{sec5}

\subsection{Caption Generation}
Figure~\ref{fig4} presents examples of image captions and final captions generated using the proposed method. 
The upper row shows the results when robot motion information was included in the VLM input, while the lower row illustrates results without motion information. Both Door opening and Bin pick-placing tasks produced appropriate captions. Notably, when robot joint angles and end-effector (EE) state information were provided, trajectory details such as direction and position were incorporated into the final captions, leading to more comprehensive descriptions. The average word count for final captions was 25.2 when motion information was excluded and 35.45 when it was included.
By inputting motion information into the VLM in multiple stages, the final captions became more stable and detailed, even when motion details were not present in the initial image captions. However, the generated captions remained consistent even without motion information input, suggesting that the VLM is capable of producing generalized or superficial outputs.

\begin{figure}[tb]
    \centering
    \includegraphics[width=\columnwidth]{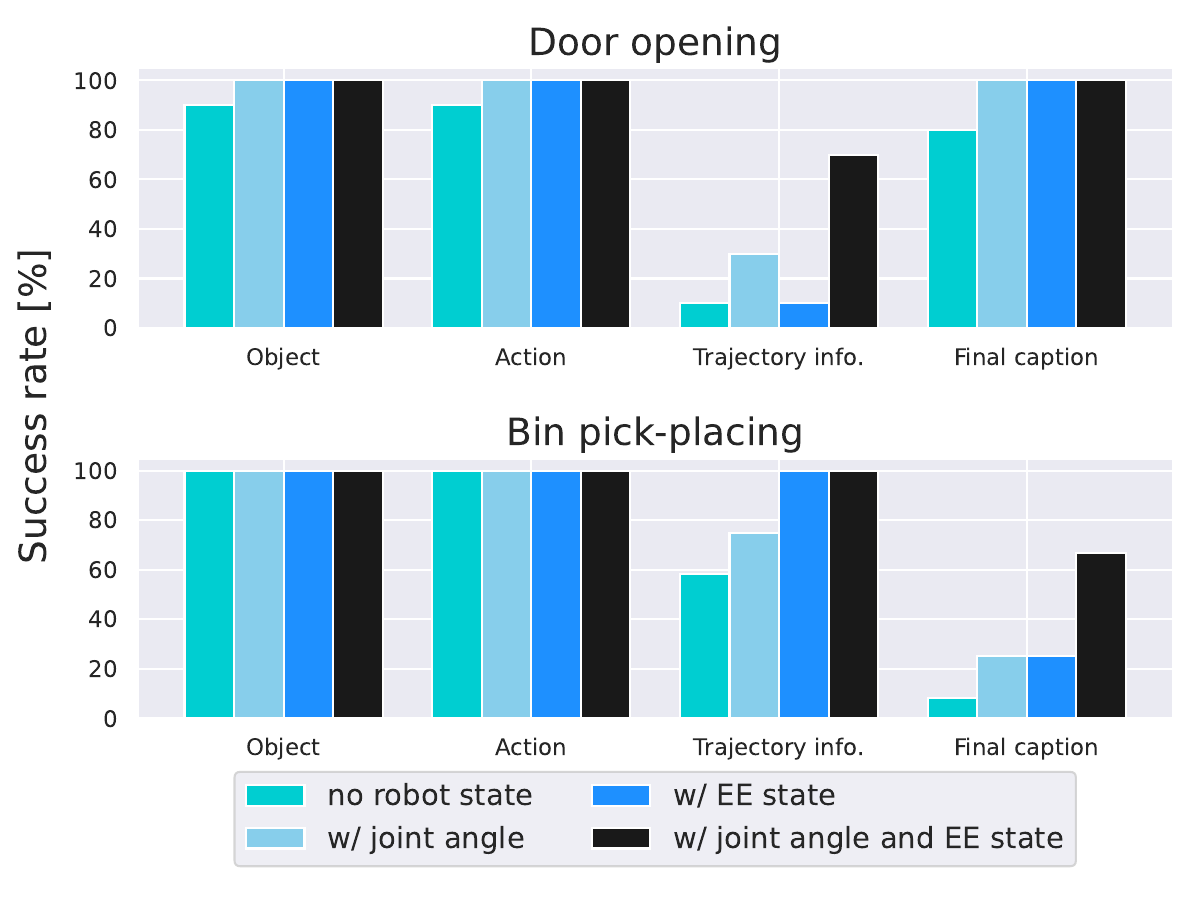}
    \caption{The success rates of caption generation task.}
    \label{fig5}
\end{figure}

\begin{figure}[tb]
    \centering
    \includegraphics[width=\columnwidth]{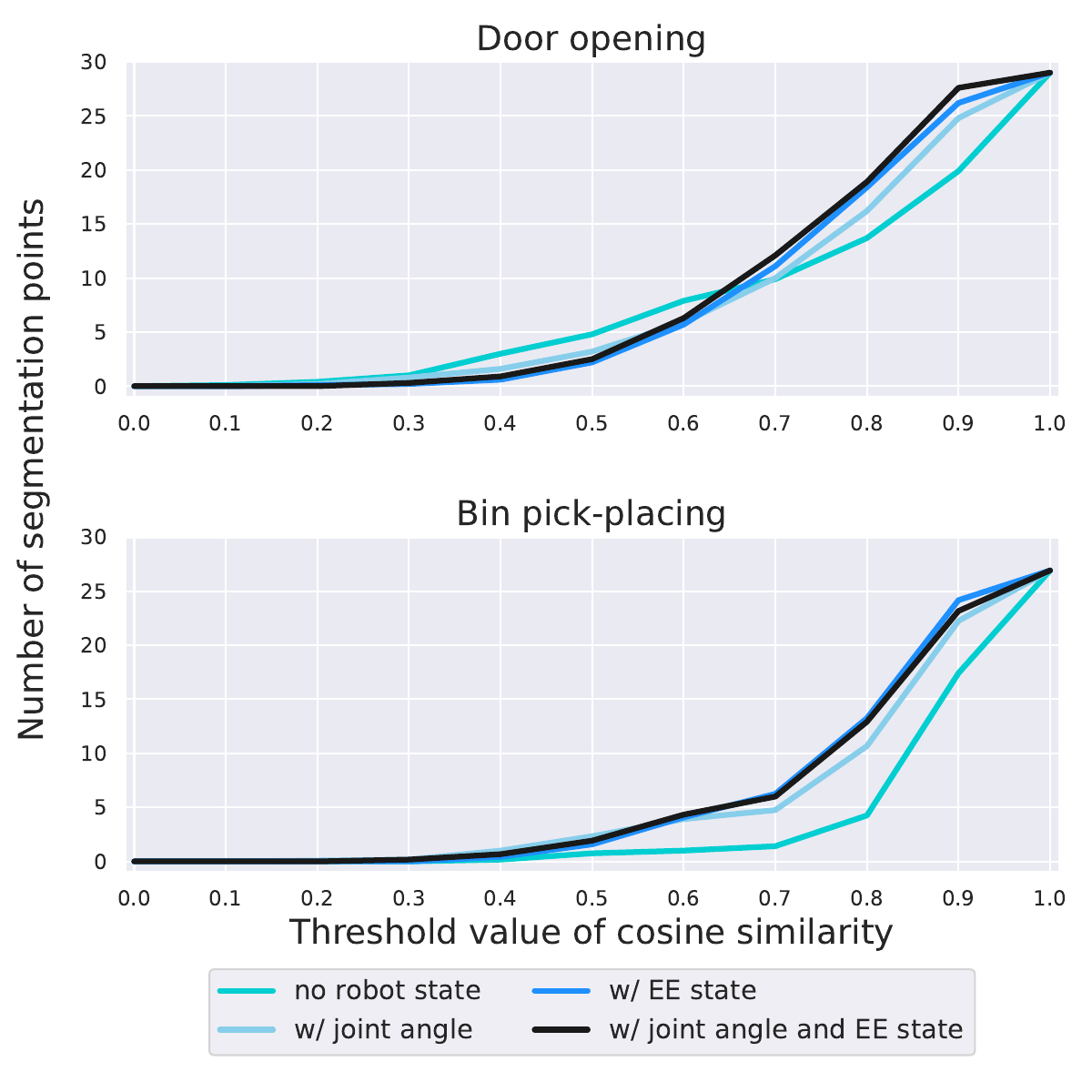}
    \caption{Number of segmentation points per threshold of cosine similarity.}
    \label{fig6}
\end{figure}

Next, we evaluated whether the final captions adequately described the target object, action, and trajectory information for each task. Figure~\ref{fig5} shows the percentage of captions that included each item (target object, action, and trajectory information) and the percentage of captions with accurate content. 
The items for the final caption were evaluated only for consistency of content.
We performed ablation experiments under four conditions (VLM w/ EE state, w/ joint angle, w/ both, and no robot state).
In the Door-opening task, no difference was observed in recognizing the target object and action, regardless of whether motion information was input into the VLM. However, including joint and end-effector states in the input enabled the generation of directional information, such as ``downward" and ``upward." A similar trend was observed in the Bin pick-placing task, where incorporating motion information resulted in identifying the four bin placement positions and improved caption accuracy. Thus, in both tasks, the inclusion of motion information enhanced the quality of the final captions. This suggests that the VLM can link directional and positional information in images with low-level motion data, potentially reducing the impact of noise elements such as image occlusion. However, even with motion information, generating accurate captions for fine-grained tasks remains challenging.

\subsection{Subtask Segmentation}
Finally, we verified that it is possible to appropriately segment motions by comparing the sentence similarity of image captions.
Fig.~\ref{fig6} shows the transition in the number of subtask segments according to the threshold of sentence similarity.
The vertical axis represents the number of subtask segments, while the horizontal axis indicates the threshold value. 
In both the Door opening and Bin pick-placing tasks, the difference in the number of subtask segments with and without robot motion information input tends to increase as the threshold value rises. 
This suggests that the trajectory information embedded in the image captions generated during the first stage of our method plays a significant role. However, it was observed that if the threshold is set too high, the segmentation points become less intuitive and harder for humans to interpret (as explained in the following paragraph). Therefore, selecting an appropriate threshold is crucial, depending on the specific requirements of the subsequent robot learning task.

\begin{figure}[tb]
    \centering
    \includegraphics[width=\columnwidth]{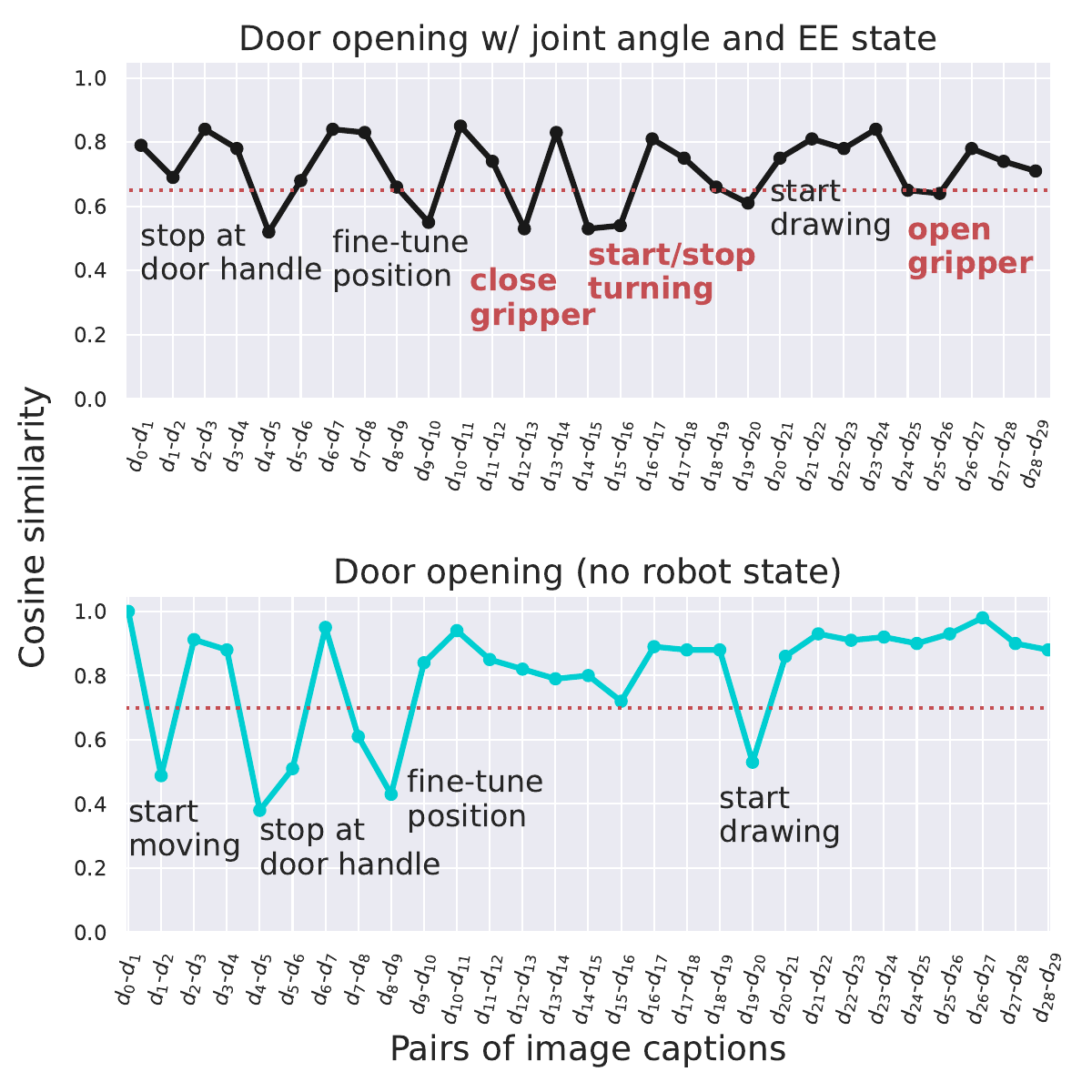}
    \caption{Examples of generated subtask segmentation (Door opening).}
    \label{fig7}
\end{figure}

\begin{figure}[tb]
    \centering
    \includegraphics[width=\columnwidth]{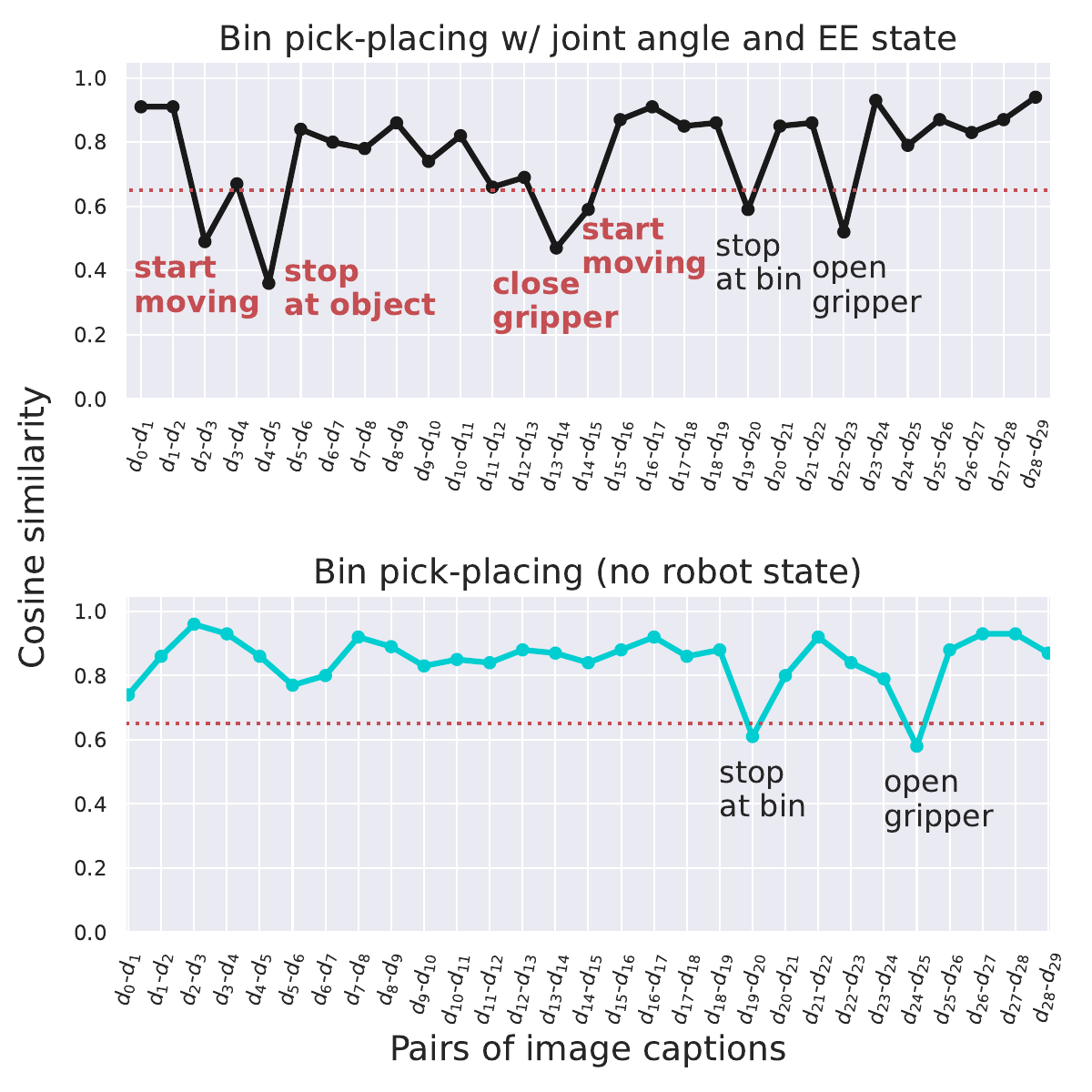}
    \caption{Examples of generated subtask segmentation (Bin pick-placing).}
    \label{fig8}
\end{figure}

Examples of subtask segmentation results for each task are shown in Figs.~\ref{fig7} and \ref{fig8}.
The vertical axis represents the cosine similarity, the horizontal axis represents the set of image captions, and the dashed line indicates the cosine similarity threshold.
Each subtask segmentation was performed with a threshold of 0.65.
Each text in the figure represents a segmentation point.
In both the Door opening and Bin pick-placing tasks, the results aligned with human intuition, with more detailed segmentation points appearing when motion information was included. 
Notably, gripper open/close data and trajectory details such as direction and position, which were reflected in the image captions, significantly influenced the subtask segmentation points (highlighted in red in Figs.~\ref{fig7} and \ref{fig8}). As previously mentioned, even at lower thresholds, the inclusion of motion information produced more accurate segmentation points, demonstrating the effectiveness of the proposed method. However, in some cases, sentence similarity was affected by the structure of the image captions, resulting in incorrect segmentation. This is because the sentence similarity calculation relies purely on linguistic information.
And, when the image information does not change significantly, the impact on the caption result is small.
These findings indicate that while the VLM demonstrates a certain level of understanding of real-world dynamics, including robot motions, it struggles to predict subtle time-series changes.

\section{CONCLUSION}
\label{sec6}

In this study, we proposed a method for automatic captioning of robot motions using a VLM that incorporates motion information into prompts and explored whether the VLM understands robot motions.
In the proposed method, the scene captions were generated from the image captions and trajectory data of a series of robot motions, and the entire task caption was generated by summarizing them.
We also divided the series of motions into subtasks by comparing the similarity between text embedding of image captions.
Experiments using two robot tasks on a simulator demonstrated the effectiveness of the proposed method.
Results confirmed that the accuracy of generated captions and subtask segmentation improved when motion information (joint angles and end-effector state) was included in the VLM input. The findings also suggested that while the VLM has a superficial understanding of robot motion, it struggles to predict fine changes over time. 
In future work, we plan to extend the proposed method by incorporating it into imitation learning frameworks~\cite{Suzuki2023} and LLM-based motion planning~\cite{Hori2023}.




\section*{ACKNOWLEDGMENT}
This work was supported by JST Moonshot R\&D Grant Number JPMJMS2031 and JSPS Grant-in-Aid for Early-Career Scientists (Grant Number: 24K20877), Japan.

\bibliographystyle{IEEEtran}
\bibliography{main}

\end{document}